\pdfoutput=1
\documentclass[11pt]{article}

\usepackage{acl}

\usepackage{times}
\usepackage{latexsym}

\usepackage[T1]{fontenc}
\usepackage[utf8]{inputenc}

\usepackage{microtype}

\usepackage{inconsolata}

\usepackage{amsmath,amssymb,amsfonts}%
\usepackage{amsthm}%
\usepackage{mathrsfs}%

\usepackage{verbatim} % for paragraphs comment
\usepackage{multirow}
\usepackage{booktabs} % table partial line
\usepackage{makecell}
\usepackage{CJK}
\usepackage{graphicx} % draw figure
\usepackage{caption} % subfigure
\usepackage{subcaption} % subfigure
\usepackage{setspace}
\usepackage{cleveref}

\title{Self-Improving for Zero-Shot Named Entity Recognition \\with Large Language Models}

\author{
	Tingyu Xie$^{1,2}$, Qi Li$^{1,2}$, Yan Zhang$^3$\thanks{ ~ Corresponding authors.} ,
	\textbf{Zuozhu Liu$^2$, Hongwei Wang$^{1,2*}$}\\
	$^1$College of Computer Science and Technology, Zhejiang University, China \\$^2$ZJU-UIUC Institute, Zhejiang University, China \\$^3$National University of Singapore, Singapore\\
	\{tingyuxie, hongweiwang\}@zju.edu.cn, yanzhang.jlu@gmail.com
}

\begin{document}
\maketitle
\begin{abstract}

Exploring the application of powerful large language models (LLMs) on the named entity recognition (NER) task has drawn much attention recently. This work pushes the performance boundary of zero-shot NER with LLMs by proposing a training-free self-improving framework, which utilizes an unlabeled corpus to stimulate the self-learning ability of LLMs. First, we use the LLM to make predictions on the unlabeled corpus using self-consistency and obtain a self-annotated dataset. Second, we explore various strategies to select reliable annotations to form a reliable self-annotated dataset. Finally, for each test input, we retrieve demonstrations from the reliable self-annotated dataset and perform inference via in-context learning. Experiments on four benchmarks show substantial performance improvements achieved by our framework. Through comprehensive experimental analysis, we find that increasing the size of unlabeled corpus or iterations of self-improving does not guarantee further improvement, but the performance might be boosted via more advanced strategies for reliable annotation selection.\footnote{Code and data are publicly available: \url{https://github.com/Emma1066/Self-Improve-Zero-Shot-NER}}
% Through comprehensive experimental analysis, our study yielded the following findings: (1) The self-improving framework further pushes the boundary of zero-shot NER with LLMs, and achieves an obvious performance improvement; (2) Iterative self-improving or naively increasing the size of unlabeled corpus does not guarantee improvements; (3) There might still be space for improvement via more advanced strategy for reliable entity selection.
% (2) there is no significant performance differences between various reliable sample selection strategies;

\end{abstract}

\section{Introduction}
% The remarkable zero-shot and few-shot generalization of large language models (LLMs) \citep{openai_blog_22,touvron2023llama,chowdhery2022palm} have brought huge revolution on various natural language processing tasks. Named entity recognition (NER) is a fundamental task in information extraction (IE). 
There have been many works exploring new possibilities of the named entity recognition (NER) task in the era of large language models (LLMs) \citep{openai_blog_22,touvron2023llama,chowdhery2022palm} recently. These studies include designing advanced prompting methods for zero-shot prediction or few-shot in-context learning (ICL) \citep{wei2023zeroshot,wang2023gptner,xie2023empirical,li-etal-2023-codeie}, training task-specific LLMs for NER \citep{zhou2023universalner,sainz2023gollie}, and generating data with LLMs to train small specific models \citep{zhang2023llmaaa,ma2023star,josifoski2023exploiting}.
% Some works study advanced prompting methods to improve zero-shot prediction or few-shot in-context learning (ICL) for NER. \citep{wang2023gptner,xie2023empirical,wei2023zeroshot}. Some works train task-specific LLMs for NER \citep{sainz2023gollie,zhou2023universalner}. Also, there are studies use LLMs as data annotator or data generator to conduct data augmentation for small language models \citep{ma2023star,josifoski2023exploiting}.

In this work, we explore the possibility of pushing the performance boundary of zero-shot NER with LLMs via self-improving. We focus on the strict zero-shot scenarios where no annotated data is available but only an unlabeled corpus is accessible, and no training resource or auxiliary models are available. We propose a totally training-free self-improving framework for NER, which utilizes an unlabeled corpus to stimulate the self-learning ability of LLMs. The framework consists of the following three steps. (1) Step 1: we use LLMs to self-annotate the unlabeled corpus using self-consistency (SC, \citealp{wang2022self}). Each annotated entity is associated with a SC score, which is used as the measure of the reliability of this annotation. (2) Step 2: we select reliable annotation to form a reliable self-annotated dataset, during which diverse annotation selection strategies are explored, including entity-level threshold filtering, sample-level threshold filtering and two-stage majority voting. (3) Step 3: for each arrived test input, we perform inference via ICL with demonstrations from the reliable self-annotated dataset. Various strategies for demonstration retrieval are explored.

\begin{figure*}[t]
	\centerline{\includegraphics[width=\linewidth]{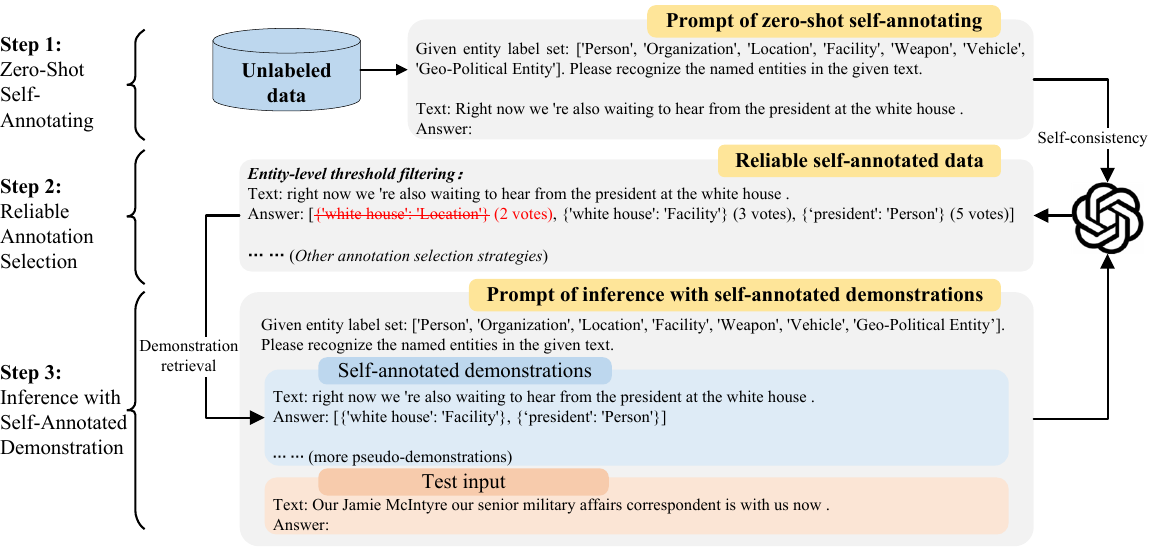}}
	\caption{The overview of the proposed self-improving framework for zero-shot NER with LLM.}
	% The framework consists of three steps: (1) \textbf{Zero-shot self-annotating}, where we use LLM to self-annotate the unlabeled corpus using self-consistency. (2) \textbf{Reliable annotation selection}, where we select reliable annotations via various strategies respectively. Here, entity-level threshold filtering with threshold set to 3 is illustrated as an example strategy.  (3) \textbf{Inference with reliable self-annotated demonstration}, where for each arrived test input, we retrieve demonstrations from the reliable self-annotated dataset and conduct ICL.
	\label{fig:method}
	\vspace{-1em}
\end{figure*}

Our contributions include: (1) We proposed a training-free self-improving framework for zero-shot NER with LLMs. (2) This framework achieved significant performance improvements on four benchmarks. (3) We conduct comprehensive experimental analysis, finding that increasing the size of unlabeled corpus or iterations of self-annotating does not guarantee gains, but there might be room for improvements with more advanced strategies for reliable annotation selection.

\section{Zero-Shot NER with Self-Improving}
\textbf{Motivation}.\quad
To push the performance boundary of zero-shot NER with LLMs, we propose a self-improving framework under a strict zero-shot and low-resource setting: No annotated data but only an unlabeled corpus is available; No auxiliary model or training step is required. This study is orthogonal to previous prompt designing works, as any advanced prompting method can be applied to this framework. Fig. \ref{fig:method} shows the framework overview.

\begin{spacing}{1}
\end{spacing}
\noindent
\textbf{Task Formulation}.\quad
Given an input sentence $x$, the NER task is to recognize the structure output $y$ from $x$, which consists of a set of $(e,t)$ pairs. $e$ is an entity span, which is a sequence of tokens form $x$; $t$ is the corresponding entity type, which belongs to a predefined entity type set.

%\subsection{Step 1: Self-Annotating Unlabeled Corpus}
\subsection{Step 1: Zero-Shot Self-Annotating}
We assume an unlabeled corpus $\mathcal{U}=\{x_i\}^n_{i=1}$ is available. We use the training set without labels as the unlabeled dataset in this work. For unlabeled sample $x_i$, we generate predictions with LLMs via zero-shot prompting, as shown in upper part of Fig. \ref{fig:method}. This process is formulated as $y_i=\mathop{\arg\max}_{y}P(y|T, x_i)$, where $T$ is the task instruction of NER, and $y_i=\{(e_i^j,t_i^j)\}_{j=1}^{m}$.
We apply self-consistency (SC) \citep{wang2022self} to obtain a SC score for each prediction, which will be used in step 2 for reliable annotation selection. We sample multiple answers from the model, and the vote for each predicted entity $(e_i^j,t_i^j)$ is the times it appeared in all the sampled answers, which we denoted as entity-level SC score $c_i^j$. Then we get the sample-level SC score $c_i$ for each input sentence $x_i$ by taking the average SC score over all predicted entities in this sentence, i.e., $c_i=\frac{1}{m}\sum_j c_i^j$.
%The self-annotated data with entity-level and sample-level SC scores are shown in the right part of Fig. \ref{fig:method}.
For each self-annotated sample with SC scores, we can denote it as $(x_i, \{(e_i^j,t_i^j, c_i^j)\}_{j=1}^{m}, c_i)$.

\begin{table*}[t]
	\renewcommand\arraystretch{0.8}
	\centering
	\small
	\begin{tabular}{l c c c c c}
		\toprule
		\textbf{Method} & \textbf{CoNLL03} & \textbf{ACE05} & \textbf{WikiGold} & \textbf{GENIA} & \textbf{Avg} \\ \midrule
		No-demos & 68.97 $_{ 0.22}$  & 27.29 $_{0.58}$ & 70.8 & 47.41 $_{0.29}$ & 53.62 \\ \midrule
		\multicolumn{6}{c}{\textbf{ICL with self-annotated demonstrations (Zero-shot)}} \\ \midrule
		\textit{Without annotation selection} & ~ & ~ & ~ & ~ & ~ \\ 
		Random & 71.45 $_{0.10}$ & 30.38 $_{0.93}$ & 70.51 & 48.78 $_{0.06}$ & 55.28 \\ 
		Nearest & 72.07 $_{0.11}$ & \textbf{32.20} $_{0.92}$ & \textbf{71.81} & 49.54 $_{1.88}$ & \textbf{56.40} \\ 
		Diverse Nearest, random & \textbf{72.15} $_{0.65}$ & 31.07 $_{1.45}$ & 70.72 & \textbf{50.01} $_{1.20}$ & 55.99 \\
		\midrule
		\textit{Entity-level threshold filtering} & ~ & ~ & ~ & ~ & ~ \\ 
		Random & 70.91 $_{0.55}$ & 30.41 $_{0.95}$ & 72.33 & 50.70 $_{1.53}$ & 56.09 \\ 
		Nearest & 73.24 $_{0.53}$ & 32.22 $_{0.38}$ & 72.53 & 49.85 $_{1.20}$ & 56.96 \\ 
		Diverse Nearest, random & 74.11 $_{0.12}$ & \underline{\textbf{32.29}} $_{0.31}$ & 72.01 & 50.68 $_{0.14}$ & 57.27 \\ 
		Diverse Nearest, SC ranking & \underline{\textbf{74.99}} $_{0.20}$ & 31.65 $_{0.97}$ & \textbf{73.53} & \textbf{51.11} $_{0.28}$ & \textbf{57.82} \\ \midrule
		\textit{Sample-level threshold filtering} & ~ & ~ & ~ & ~ & ~ \\ 
		Random & 72.41 $_{1.28}$ & 30.00 $_{1.26}$ & \textbf{73.38} & 51.61 $_{1.21}$ & 56.86 \\ 
		Nearest & 72.28 $_{0.14}$ & \textbf{32.00} $_{0.08}$ & 73.27 & \underline{\textbf{52.72}} $_{0.80}$ & \textbf{57.57} \\ 
		Diverse Nearest, random & 72.32 $_{0.08}$ & 30.74 $_{0.06}$ & 72.09 & 52.50 $_{0.50}$ & 56.91 \\ 
		Diverse Nearest, SC ranking & \textbf{73.97} $_{0.12}$ & 31.08 $_{0.54}$ & 72.80 & 51.67 $_{0.93}$ & 57.38 \\ \midrule
		\textit{Two-stage majority voting} & ~ & ~ & ~ & ~ & ~ \\ 
		Random & 72.12 $_{0.59}$ & 31.18 $_{0.38}$ & 72.32 & 50.17 $_{0.93}$ & 56.45 \\ 
		Nearest & 71.66 $_{0.37}$ & 31.45 $_{1.32}$ & 72.84 & 50.19 $_{1.59}$ & 56.53 \\ 
		Diverse Nearest, random & 72.45 $_{0.41}$ & 30.84 $_{0.56}$ & 70.83 & 51.03 $_{0.73}$ & 56.28 \\ 
		Diverse Nearest, SC ranking & \textbf{74.51} $_{0.03}$ & \textbf{32.27} $_{0.25}$ & \underline{\textbf{73.98}} & \textbf{52.06} $_{0.09}$ & \underline{\textbf{58.20}}\\ \midrule
		\multicolumn{6}{c}{\textbf{ICL with gold labeled demonstrations}} \\ \midrule
		Random (Gold) & 78.36 $_{0.31}$ & 42.12 $_{0.30}$ & 74.27 & 54.50 $_{1.14}$ & 62.31 \\ 
		Nearest (Gold) & 84.30 $_{0.39}$ & 52.72 $_{0.44}$ & 78.20 & 54.78 $_{0.94}$ & 67.50 \\ \midrule
		Random (Gold), full data & 78.35 $_{1.44}$ & 41.33 $_{0.79}$ & 78.47 & 52.77 $_{2.03}$ & 62.73 \\ 
		Nearest (Gold), full data & 83.51 $_{0.02}$ & 55.54 $_{0.61}$ & 79.73 & 58.72 $_{1.52}$ & 69.37 \\ \bottomrule
	\end{tabular}
	\caption{Main results. The right subscript number are standard deviations. \textit{Gold} indicates the method has access to the gold labeled data, thus is not comparable with the rest of methods. \textit{Full data} indicates the method has access to the full training set. Results of $Th\_entity=4.0$ and $Th\_sample=4.0$ is shown here. Texts in \textbf{bold} are the best results in each category; Text \underline{underlined} are the best results among all methods. The proposed framework significantly improves the zero-shot performances. On average, two-stage majority voting combined with the proposed diverse nearest with SC ranking achieves the best results.}
	% Results of the proposed self-improving framework with different reliable annotation selection strategies and demonstration retrieval methods.
	\label{tab:main_results}
	\vspace{-1em}
\end{table*}

\subsection{Step 2: Reliable Annotation Selection}

% We select high-quality annotation according to the entity-level SC score $c_i^j$ and sample-level SC score $c_i$ respectively to get a reliable self-annotated dataset. We assume a higher SC score indicates a higher reliability. We investigate the three following strategies for reliable annotation selection.

We assume that a higher SC score indicates a higher reliablity. Thus, we investigate the three following strategies for reliable annotation selection. (1) \textit{Entity-level threshold filtering}, which drops the predicted entity $e_i^j$ if $c_i^j<Th\_entity$, where $Th\_entity$ is the threshold for entity-level SC score. (2) \textit{Sample-level threshold filtering}, which drops the sample $x_i$ if $c_i<Th\_sample$, where $Th\_sample$ is the threshold for sample-level SC score. (3) \textit{Two-stage majority voting} \citep{xie2023empirical}, is an entity-level selection method, which first votes for the most consistent entity spans, then the most consistent types based on the voted spans.

\begin{comment}
\begin{spacing}{1.5}
\end{spacing}
\noindent
\textbf{Entity-level threshold filtering}, which drops the predicted entity $e_i^j$ if $c_i^j<Th\_entity$, where $Th\_entity$ is the threshold for entity-level SC score.
\begin{spacing}{1.5}
\end{spacing}
\noindent
\textbf{Sample-level threshold filtering}, which drops the sample $x_i$ if $c_i<Th\_sample$, where $Th\_sample$ is the threshold for sample-level SC score.
\begin{spacing}{1.5}
\end{spacing}
\noindent
\textbf{Two-stage majority voting}, proposed in \citep{xie2023empirical}, is a selection strategy on the entity-level, which first votes for the most consistent entity spans, then the most consistent types based on the voted spans.
\end{comment}

% We do not assume any specialized pre-trained models for NER to simulate the strict low-resource scenario. If a pre-trained NER model is already available, we can directly using the NER model to perform recognition instead of using the LLM. We only assume a universal sentence embedding model is available. As this kind of model could be trained on corpus from any task.

\subsection{Step 3: Inference with Self-Annotated Demonstration}
When a test input $x^q$ arrives, we retrieve $k$ demonstrations from the reliable self-annotated dataset to help the inference. \footnote{Different from \citet{lyu-etal-2023-z}, our demonstrations are obtained through self-annotating with LLMs instead of randomly assignment. Besides, randomly assigning label is not feasible for NER task as it naturally requires label information on each token. } We investigate the following four methods for demonstration retrieval. (1) \textit{Random retrieval}, which randomly select $k$ demonstrations. (2) \textit{Nearest retrieval}, which select the $k$ nearest neighbors of $x^q$. The distance of samples is measured by the cosine similarity in the representation space.
% We first obtain the representations of all samples, then we retrieve $k$ nearest neighbors in the representation space. Cosine similarity is taken to measure the distances in the representation space. 
(3) \textit{Diverse nearest retrieval}, which first retrieve $K$ nearest neighbors, where $K>k$, then uniformly samples a random set of $k$ samples from the $K$ neighbors. (4) \textit{Diverse nearest with SC ranking}, proposed by this work to achieve a better trade-off between the similarity, diversity and reliability of self-annotated demonstrations. After retrieving $K$ nearest neighbors, we select samples with the top-$k$ sample-level SC scores.
% it ranks them with sample-level SC scores, then select samples with the top-$k$ scores.

Let $S=\{x_i,y_i\}^k_{i=1}$ denotes the self-annotated demonstrations retrieved for the test input $x^q$. Finally, our framework conduct ICL by concatenating these $k$ samples as well as the test input sentence $x^q$, as shown in the below part in Fig. \ref{fig:method}. The prediction is obtained via $y^q=\mathop{\arg\max}_{y} P(y|T,S,x^q)$.

\section{Experiment}
\subsection{Setup}
We experiment on four widely-used NER datasets, CoNLL03 \citep{sang2003introduction}, ACE05 \citep{walker2006ace}, WikiGold \citep{balasuriya2009named} and GENIA \citep{ohta2002genia}. We use GPT3.5 (gpt-3.5-turbo) as the LLM backbone and text-embedding-ada-002 model to get sentence representations.\footnote{The results of GPT-3.5 are obtained during October and November 2023 with official API.} We set $k=16$ and $K=50$. For SC, we set temperature to 0.7 and sample 5 answers. For cost saving, we randomly sample 300 test samples twice then report the means and standard deviations, and we randomly sample 500 training samples without labels to form the unlabeled corpus $\mathcal{U}$. The naive zero-shot prompting is our baseline, which we denote as \textit{No-demos}. We report F1 scores throughout this paper.

\subsection{Results}
\label{sec:main_results}
The main results are shown in Table \ref{tab:main_results}. Results of other values for thresholds $Th\_entity$ and $Th\_sample$ can be found in Appendix \ref{sec:app_various_th}. (1) Without annotation selection, we only generate one answer for each unlabeled sample. The results show improvements over \textit{No-demos}, revealing that our framework is helpful even without any carefully designed annotation selection step. 
%In the next three parts, we explore the effects of different strategies for reliable sample selection.
(2) The performance is further improved under three annotation selection strategies respectively.
(3) The proposed diverse nearest with SC ranking shows consistent improvements under various settings and achieves the best results when combined with two-stage majority voting. This confirms that this strategy achieves a better trade-off between similarity, diversity and reliability of the demonstrations.
(4) Random retrieval lags behind nearest retrieval in self-improving scenario but is not as much as in the gold label scenario, likely because of the noise contained in self-annotated labels. The model may directly copy the wrong answers in the most similar self-annotated demonstrations due to the copy mechanism of ICL \citep{lyu-etal-2023-z}.

\begin{comment}
\begin{figure}[t]
	\centering
	\begin{subfigure}{0.35\linewidth}
		\centering
		\includegraphics[width=\linewidth]{figs/ulb_size_conll.pdf}
		\caption{CoNLL03}
		\label{fig:ulb_size_conll}
	\end{subfigure}
	\begin{subfigure}{0.35\linewidth}
		\centering
		\includegraphics[width=\linewidth]{figs/ulb_size_ace05.pdf}
		\caption{ACE05}
		\label{fig:ulb_size_ace05}
	\end{subfigure}
	\begin{subfigure}{0.35\linewidth}
		\centering
		\includegraphics[width=\linewidth]{figs/ulb_size_wiki.pdf}
		\caption{WikiGold}
		\label{fig:ulb_size_wiki}
	\end{subfigure}
	\begin{subfigure}{0.35\linewidth}
		\centering
		\includegraphics[width=\linewidth]{figs/ulb_size_genia.pdf}
		\caption{GENIA}
		\label{fig:ulb_size_genia}
	\end{subfigure}
	\caption{Increasing the size of unlabeled dataset does not guarantee performance gains.}
	\label{fig:ulb_size_conll_wiki}
	\vspace{-1.3em}
\end{figure}
\end{comment}

\begin{figure}[t]
	\centerline{\includegraphics[width=\linewidth]{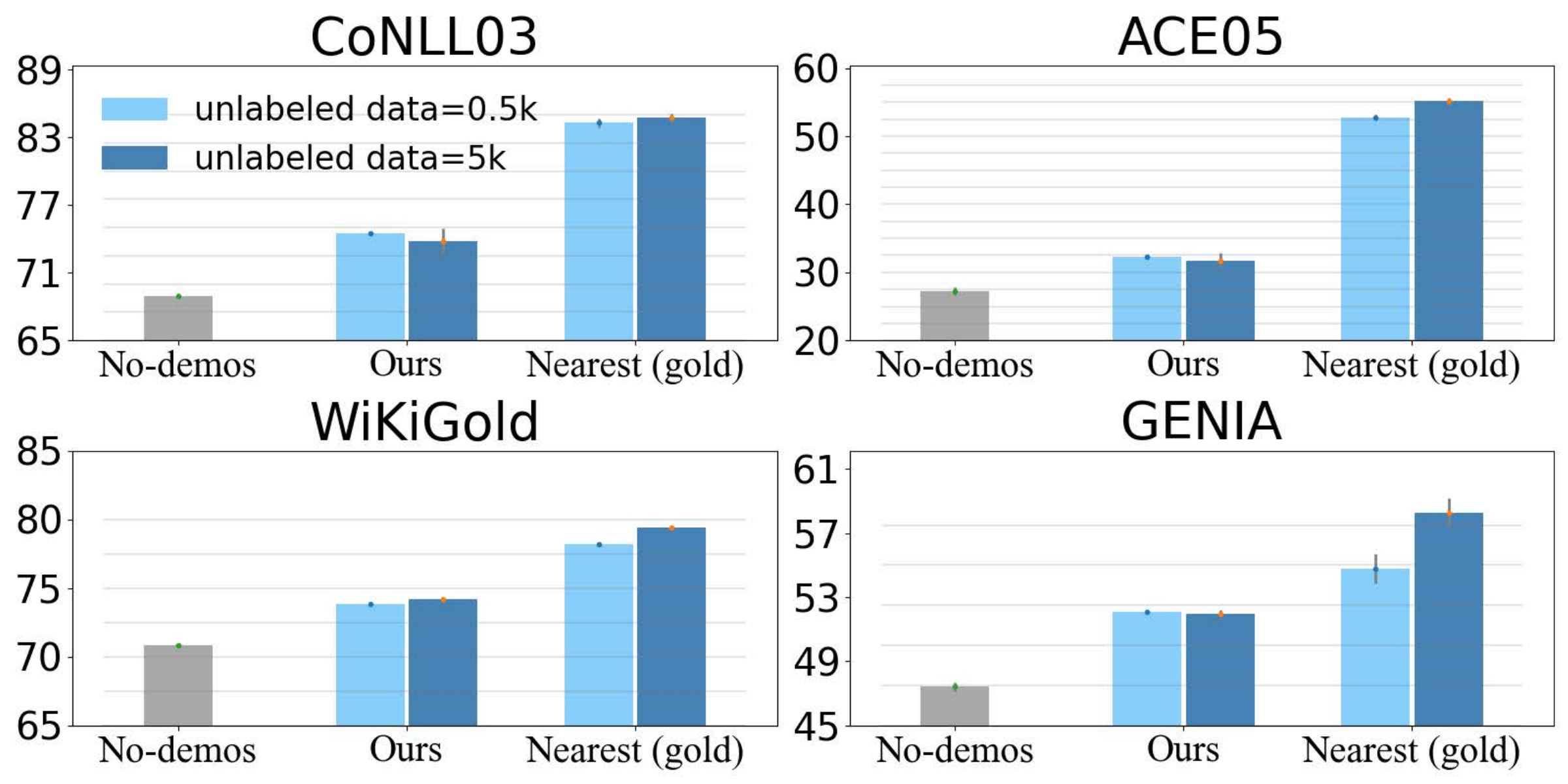}}
	\caption{Results of increasing the size of unlabeled dataset. Vertical axes represent F1 scores. \textit{Ours} refers to the combination of two-stage majority voting and diverse nearest with SC ranking. Increasing unlabeled data does not guarantee performance gains.}
	\label{fig:ulb_size_all}
	\vspace{-0.5em}
\end{figure}

\begin{figure}[t]
	\centerline{\includegraphics[width=0.8\linewidth]{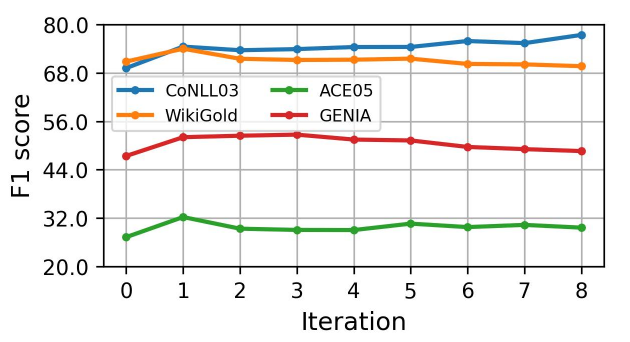}}
	\caption{Increasing the iterations of self-improving does not guarantee performance improvements.}
	\label{fig:iterations_all}
	\vspace{-0.5em}
\end{figure}

\subsection{Analysis}
\label{sec:analysis}

\begin{figure}[t]
	\centerline{\includegraphics[width=\linewidth]{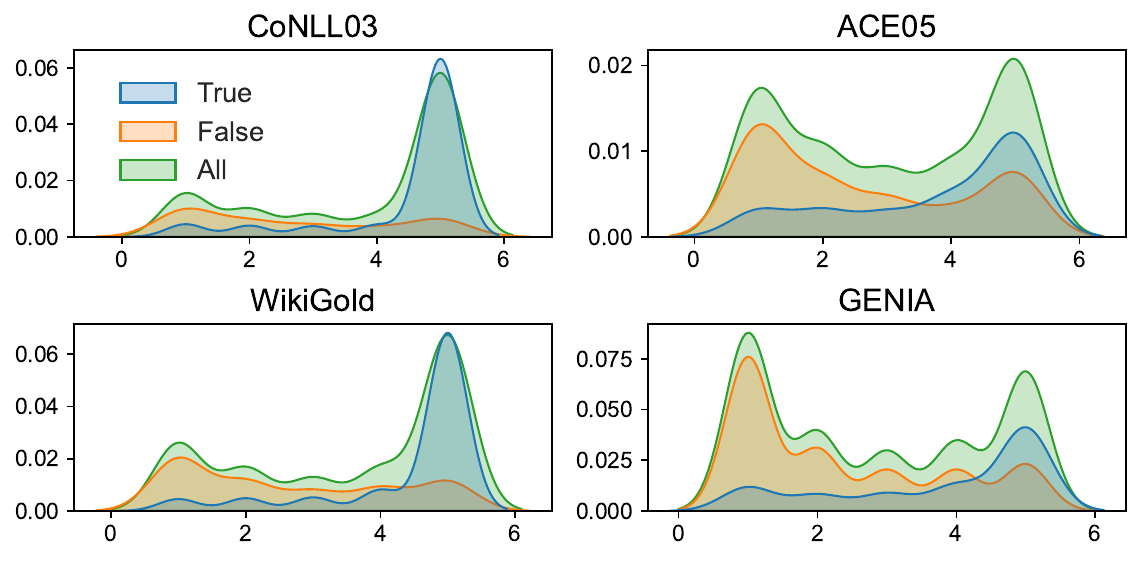}}
	\caption{Kernel density estimation for SC scores. Vertical axes represent density, horizontal axes represent SC scores.}
	\label{fig:kernel_all}
	\vspace{-0.5em}
\end{figure}

\begin{table}[t]
	\renewcommand\arraystretch{0.6}
	\centering
	\small
	\tabcolsep=0.2em
	\begin{tabular}{l c c c c c}
		\toprule
		\textbf{Method} & \textbf{CoNLL03} & \textbf{ACE05} & \textbf{WikiGold} & \textbf{GENIA} & \textbf{Avg} \\ \midrule
		No-demos & 68.97 & 27.29 & 70.8 & 47.41 & 53.62 \\ \midrule
		TSMV & 74.51 & 32.27 & 73.98 & 52.06 & 58.20 \\ 
		Upper bound & 81.65 & 37.82 & 76.57 & 56.24 & 63.07 \\ \midrule
		Gold label & 84.30 & 52.72 & 78.20 & 54.78 & 67.50 \\ \bottomrule
	\end{tabular}
	\caption{Results of the upper bound of reliable annotation selection. \textit{TSMV} represents two-stage majority voting. We display the best results for each strategy. The setting of \textit{Upper bound} performs on par with the setting of \textit{Gold label}, showing that there might be space to be improved for reliable annotation selection.}
	\label{tab:SC_analy_results}
	\vspace{-0.5em}
\end{table}

\noindent
\textbf{Increasing unlabeled data}.
% We explore the effect of increasing the size of unlabeled corpus $\mathcal{U}$. 
We expanded the size of $\mathcal{U}$ by 10 times and randomly sampled 5000 samples from the original training set. Results are shown in Fig. \ref{fig:ulb_size_all}. Increasing the size of the unlabeled corpus does not guarantee performance improvements under the self-improving scenario. Meanwhile, increasing the size of the demonstration pool only brings marginal improvement, even under the gold label scenario. The reason may be that the small dataset already approximately captures the data distribution.
% This observation is consistent with the results on the low-resource scenario in \citep{wang2023gptner}. 
% increasing data size cannot bring obvious improvements. 
% 

\begin{spacing}{1.2}
\end{spacing}
\noindent
\textbf{Iterative self-improving}.
% We explore the effect of iterative self-improving: 
We use the self-annotated data as demonstrations to guide the next iteration of self-annotating, forming a bootstrapping process. The illustration of iterative self-improving process can be found in Appendix \ref{sec:app_iterations}. We experiment up to 8 iterations. The 0-th iteration indicates the \textit{No-demos} setting. Results are shown in Fig.  \ref{fig:iterations_all}. Increasing iterations of self-improving cannot guarantee improvements on most datasets. 
% The performance generally keeps climbing with the increase of iterations on CoNLL03. However, the performances keeps dropping or wandering around the original level on other datasets. 
This may due to the fact that error accumulation in self-annotating is difficult to be eliminated in this training-free process.

\begin{spacing}{1.2}
\end{spacing}
\noindent
\textbf{Upper bound of reliable annotation selection.} We keep only the true predictions and discard the false predictions in all the sampled answers to evaluate the upper bound of reliable annotation selection. Results are shown in Table \ref{tab:SC_analy_results}. More detailed results can be found in Appendix \ref{sec:app_SC_analy}. \textit{Upper bound} setting performs on par with the \textit{Gold label} setting, indicating that there might still be space to be improved for reliable annotation selection.
% More advanced strategies can be explored for selecting reliable predictions.

\begin{spacing}{1.2}
\end{spacing}
\noindent
\textbf{SC score analysis.}
We plot the kernel density estimation for entity-level SC scores in Fig. \ref{fig:kernel_all}. Most true predictions gather in the interval of high SC scores, while most false predictions have low SC scores. This shows that SC scores effectively reflect the reliability of annotations.

\begin{table}[th]
	\centering
	\small
	\renewcommand\arraystretch{0.8}
	\tabcolsep=0.6em
	\begin{tabular}{l c c}
		\midrule
	\textbf{Method} & \textbf{SC} & \textbf{SV}  \\
		\midrule
		No-demos & \multicolumn{2}{c}{68.97 $_{0.22}$} \\
		\midrule
		\multicolumn{3}{l}{\textit{Entity-level threshold filtering}}  \\ 
		Random & 70.91 $_{0.55}$ & 70.91 $_{0.56}$  \\ 
		Nearest & 73.24 $_{0.53}$ & 71.23 $_{0.01}$ \\ 
		Diverse Nearest, random & 74.11 $_{0.12}$ & \underline{\textbf{71.44}} $_{0.93}$   \\ 
		Diverse Nearest, score ranking & \underline{\textbf{74.99}} $_{0.20}$ & 68.09 $_{0.60}$   \\ 
		\midrule
		\multicolumn{3}{l}{\textit{Sample-level threshold filtering}} \\ 
		Random & 72.41 $_{1.28}$ & \textbf{71.00} $_{0.32}$  \\ 
		Nearest & 72.28 $_{0.14}$ & 70.45 $_{0.46}$  \\ 
		Diverse Nearest, random & 72.32 $_{0.08}$ & 70.06 $_{1.29}$ \\ 
		Diverse Nearest, score ranking & \textbf{73.97} $_{0.12}$ & 68.95 $_{0.35}$ \\ 
		\bottomrule
	\end{tabular}
	\caption{Comparison between SC and SV on CoNLL03 dataset. $Th\_entity=4.0$ and $Th\_sample=4.0$ is used. Right subscript number are standard deviations. Texts in \textbf{bold} are the best results in each category; Text \underline{underlined} are the best results among all methods.}
\label{tab:sv_compare}
\vspace{-0.5em}
\end{table}

\begin{spacing}{1.2}
\end{spacing}
\noindent
\textbf{Self-verification.}
Besides SC, we also explore self-verification (SV) to measure the confidence of self-annotation by asking the LLM to score its own answer about its own confidence. After the LLM outputs the recognized entities, we obtain the SV score by asking the LLM: "How confident are you in providing the above answers? Please give each named entity in your answer a confidence score of 0-5." The comparison results between SC and SV are in Table \ref{tab:sv_compare}. As shown in the table, SV also achieves some improvements compared with the No-demos baseline. However, it lags behind the SC measurement. This is presumably because the LLM tends to be over-confident about its own answer, since we found that no sample gets a confidence score lower than 3 under the SV measurement in CoNLL03 benchmark. The overconfidence problem is also mentioned in \citet{li2023evaluating}. 

\begin{table}[t]
	\renewcommand\arraystretch{0.8}
	\centering
	\small
	\begin{tabular}{l c c}
		\toprule
		\textbf{Method} & \textbf{CoNLL03} & \textbf{WikiGold} \\
		\midrule
		No-demos & 42.24 & 28.57  \\ \midrule
		Nearest & 23.55 & 8.94  \\ \bottomrule
	\end{tabular}
	\caption{Results on the Llama2 chat 13B. Two-stage majority voting is used here. The negative results show that the proposed framework is more suitable for models with a strong zero-shot capability. The negative effect is obvious on the first sampled test set, thus we do not continue to test on other seeds.}
	\label{tab:llama2_13b}
	\vspace{-1em}
\end{table}

\begin{spacing}{1.2}
\end{spacing}
\noindent
\textbf{Evaluation on weaker LLMs.}
To explore the performance of the proposed self-improving framework on weaker LLMs, we conduct experiments on the Llama2 chat 13B model \citep{touvron2023llama},\footnote{\url{https://huggingface.co/meta-llama/Llama-2-13b-chat-hf}} the results are shown in Table \ref{tab:llama2_13b}. Two-stage majority voting selection strategy and the nearest neighbor retrieval method are used in this experiment. With a much weaker ability in zero-shot scenarios, Llama2 13B model shows negative results under the self-improving framework. This indicates that the proposed framework is more suitable for models with a strong zero-shot capability. For the models with a relatively weaker zero-shot ability, improving the prompt designing might be a more effective strategy to boost performance.

\section{Related Work}
\label{sec:app_related_work}
%[GPT-NER] design a prompt paradigm for NER by inserting special tokens around entities, and utilize the full training set as demonstrations to mitigate the performance gap between LLM and specialized fine-tuned NER models. 
%
%However, we argue that, with the full training set available, training a specialized model can obtain better performance on NER. The necessity for using LLM to conduct NER is trivial. We assume a strict scenario where no annotated data is available.
%
%[LLMAAA] also use ChatGPT for data annotation. However, they use a small model to select the annotated-data in the meantime. In our setting, we keep a training free paradigm.......
\noindent
\textbf{Information extraction with LLM.} \quad
% Some works designed advanced prompting methods to improve the performance of information extraction (IE) with LLMs \citep{wei2023zeroshot,wang2023gptner,xie2023empirical,li-etal-2023-codeie}. 
% \citet{wei2023zeroshot} propose a two-stage chatting paradigm for IE, which first recognize the types of elements then extract the mentions corresponding to each type. \citet{wang2023gptner} apply in-context learning (ICL) to NER by inserting special tokens into the demonstrations retrieved from the full gold training set. \citet{xie2023empirical} propose decomposed-QA and syntactic augmentation to boost zero-shot NER. \citet{li-etal-2023-codeie} transfer IE into code generation task.
% Some works trained task-specific LLMs for IE with instruction-tuning \citep{zhou2023universalner,sainz2023gollie}. In addition, several works used LLMs to annotate data or generate synthetic data for small expert IE models \citep{zhang2023llmaaa,ma2023star,josifoski2023exploiting}. For example, \citet{zhang2023llmaaa} use LLM to annotate data, which is used to fine-tune a small expert IE model, then use the fine-tuned model to help select the data to be annotated; this pipeline is iteratively conducted to form a bootstrapping process. 
% \citet{ma2023star,josifoski2023exploiting} generate synthetic data for IE via a structure-to-text paradigm. 
The research of information extraction (IE) with LLMs includes prompt designing \citep{wei2023zeroshot,wang2023gptner,xie2023empirical,li-etal-2023-codeie}, task-specific LLMs instruction-tuning \citep{zhou2023universalner,sainz2023gollie} and data augmentation \citep{zhang2023llmaaa,ma2023star,josifoski2023exploiting}. \citet{zhang2023llmaaa} use LLM to annotate data, which is used to fine-tune a specific IE model, then the fine-tuned model is used to help select the data to be annotated in the next iteration. Unlike previous works, this work propose a training-free self-improving framework to push the zero-shot boundary of LLM on NER. Different from \citet{zhang2023llmaaa}, no seed labeled data, expert small model nor training resources are required in our framework. 
In addition, our work is \textbf{orthogonal} to previous prompt designing works. They explored various advanced prompt formats to boost performance, and did not utilize unlabeled corpus. Unlike them, this work improves zero-shot NER by using unlabeled corpus without designing any complex prompt format. 
% Also, our work is orthogonal to the previous prompt designing works, as any advanced prompting method can be applied in the proposed framework to boost performance.

\begin{spacing}{1.2}
\end{spacing}
\noindent
\textbf{Demonstrations in ICL.} \quad
Some works explored factors that have impacts on ICL \citep{lyu-etal-2023-z,min-etal-2022-rethinking,wei2023larger}. \citet{lyu-etal-2023-z} investigate the impact of randomly assigning labels to demonstrations in ICL. However, this random labeling method is not suitable for tasks like NER, which requires label information on the token-level instead of sentence-level. Different from them, we first use LLM to make predictions on the unlabeled corpus, then select reliable self-annotated data as demonstrations.

\section{Conclusion}
We propose a training-free self-improving framework for zero-shot NER with LLMs, which achieves significant performance improvements on four benchmarks. Comprehensive experimental analysis shows that, simply increasing the size of unlabeled corpus or the iterations of self-annotation do not guarantee further improvement, but there might still be room for improvement with more advanced strategies for reliable annotation selection.

\section*{Limitations}
We acknowledge the following limitations of this study.
\begin{itemize}
		\setlength{\itemsep}{0pt}
\item This work focus on exploring the zero-shot self-improving framework on NER task. The investigation of this paradigm on other IE tasks are not studied yet. \item We explored the commonly-used self-consistency and the self-verification method to obtain the confidence score for measuring the quality of self-annotated data. There might be other approaches to measure the quality of self-annotation. 
\item The zero-shot performance still lag behind previous state-of-the-art of fully-supervised methods. 
\item Although this framework achieves significant improvement on the strong LLM, GPT-3.5, it gets negative results on a much weaker LLM, Llama2 13B. Improving the zero-shot NER on the weaker and smaller LLMs remains to be explored.
\end{itemize}
%(1) This work focus on exploring the zero-shot self-improving framework on NER task. The investigation of this paradigm on other IE tasks are not studied yet. (2) We explored the commonly-used self-consistency and the self-verification method to obtain the confidence score for measuring the quality of self-annotated data. There might be other approaches to measure the quality of self-annotation. (3) The zero-shot performance still lag behind previous state-of-the-art of fully-supervised methods. (4) Although this framework achieves significant improvement on the strong LLM, GPT-3.5, it gets negative results on a much weaker LLM, Llama2 13B. Improving the zero-shot NER on the weaker and smaller LLMs remains to be explored.

\section*{Acknowledgements}
This research is supported by Zhejiang Provincial Natural Science Foundation of China (LDT23F02023F02). We would like to thank the anonymous reviewers for their insightful comments and constructive suggestions. We would also like to thank Chen Wang and Xinlong Qiao for their help at the visualization.

\bibliography{custom}

\newpage

\appendix

\section{Dataset Statistics}
\label{sec:app_datasets}

We evaluate on four commonly-used NER English datasets, CoNLL03 \citep{sang2003introduction}, ACE05 \citep{walker2006ace}, WikiGold \citep{balasuriya2009named} and GENIA \citep{ohta2002genia}, among which CoNLL03, WikiGold and GENIA are public datasets, and ACE05\footnote{\url{https://catalog.ldc.upenn.edu/LDC2006T06}} can be accessed on Linguistic Data Consortium (LDC) platform with specific license. In addition, we also evaluate on two Chinese datasets, Ontonotes 4\footnote{\url{https://catalog.ldc.upenn.edu/LDC2011T03}} and MSRA \citep{zhang-etal-2006-word}, in Appendix \ref{sec:app_zh_benchmarks}.
Table \ref{tab:app_datasets_en} and \ref{tab:app_datasets_zh} shows the statistics of the processed datasets used in this work. For CoNLL03, we use the processed version shared by \citet{han2023information}. For ACE05, we follow \citet{luan-etal-2019-general}'s processing steps.

\begin{table}[h]
	\centering
	\small
	\tabcolsep=0.6em
	\begin{tabular}{l c c c c}
		\toprule
		\textbf{Dataset} & \textbf{CoNLL03} & \textbf{ACE05} & \textbf{WikiGold} & \textbf{GENIA}  \\ \midrule
		\#Train & 14382 & 12475 & 1422 & 16692 \\ 
		\#Test & 3453 & 2050 & 274 & 1854 \\
		\bottomrule
	\end{tabular}
	\caption{Statistics of the processed English datasets used in this work. The training set is formed by combining the original training split and development split.}
	\label{tab:app_datasets_en}
	\vspace{-1em}
\end{table}

\begin{table}[h]
	\centering
	\small
	\begin{tabular}{l c c}
		\toprule
	\textbf{Dataset} & \textbf{Ontonotes 4} & \textbf{MSRA} \\ \midrule
		\#Train & 20025 & 46364 \\ 
		\#Test  & 4346 & 4365 \\
		\bottomrule
	\end{tabular}
	\caption{Statistics of the processed Chinese datasets used in this work. The training set is formed by combining the original training split and development split.}
	\label{tab:app_datasets_zh}
	\vspace{-1em}
\end{table}

\section{Results on Additional Benchmarks}
\label{sec:app_zh_benchmarks}
We additionally evaluate on two widely-used Chinese benchmarks, the results are in Table \ref{tab:app_eval_zh}.

\begin{table}[h]
	\centering
	\small
	\tabcolsep=0.3em
	\begin{tabular}{l c c}
		\toprule
		\textbf{Method} & \textbf{Ontonotes 4} & \textbf{MSRA} \\ \midrule
		No-demos & 31.71 $_{1.14}$ & 39.21 $_{0.93}$ \\ \midrule
		\multicolumn{3}{l}{\textbf{ICL with self-annotated demonstrations}}  \\ 
		Random & 32.45 $_{0.19}$ & 39.55 $_{0.75}$ \\ 
		Nearest & 31.54 $_{1.60}$ & 36.31 $_{1.76}$ \\ 
		Diverse Nearest, SC ranking & \textbf{35.57} $_{1.22}$ & \textbf{40.84} $_{2.83}$ \\ 
		\midrule
		\multicolumn{3}{l}{\textbf{ICL with gold labeled demonstrations}} \\ 
		Random (Gold) & 49.42 $_{0.22}$ & 53.51 $_{1.38}$ \\ 
		Nearest (Gold) & 64.16 $_{1.08}$ & 61.58 $_{1.58}$ \\
		\bottomrule
	\end{tabular}
\caption{Results on Chinese benchmarks. Right subscript numbers are standard deviations. \textit{Gold} indicates access to the gold labeled data, thus is not comparable with the rest of methods. Two-stage majority voting is used here. Texts in \textbf{bold} are the best results.}
\label{tab:app_eval_zh}
	\vspace{-1em}
\end{table}

\section{Results on Other Embedding Models}
\label{sec:app_embs}
We explore the effect of using other embedding models for retrieval, SBERT \citep{reimers-gurevych-2019-sentence}\footnote{\url{https://huggingface.co/sentence-transformers/all-mpnet-base-v2}} and GTE \citep{li2023towards}\footnote{\url{https://huggingface.co/thenlper/gte-large}}. Results are in Table \ref{tab:app_embs}.

\begin{table*}[t]
	\centering
	\small
	\begin{tabular}{l c c c c c c}
		\toprule
		\textbf{Datasets} & \multicolumn{3}{c}{\textbf{CoNLL2003}} & \multicolumn{3}{c}{\textbf{WikiGold}} \\ 
		\midrule
		Embedding Models & embed-ada & SBERT & GTE & embed-ada & SBERT & GTE \\ 
		\midrule
		No-demos & 68.97 $_{0.22}$ & 68.97 $_{0.22}$ & 68.97 $_{0.22}$ & 70.80 & 70.80 & 70.80 \\ \midrule
		\multicolumn{7}{c}{\textbf{ICL with self-annotated demonstrations (Zero-shot)}} \\ \midrule
		Random & 72.12 $_{0.59}$ & 72.12 $_{0.59}$ & 72.12 $_{0.59}$ & 72.32 & 72.32 & 72.32 \\ 
		Nearest & 71.66 $_{0.37}$ & 72.07 $_{0.22}$ & 72.37 $_{1.17}$ & 72.84 & 72.39 & 72.24 \\ 
		Diverse Nearest, SC ranking & \textbf{74.51} $_{0.03}$ & \textbf{72.67} $_{0.37}$ & \textbf{72.53} $_{0.96}$ & \textbf{73.98} & \textbf{76.08} & \textbf{73.60} \\ \midrule
		\multicolumn{7}{c}{\textbf{ICL with gold labeled demonstrations}} \\ \midrule
		Random (Gold) & 77.25 $_{1.39}$ & 77.25 $_{1.39}$ & 77.25 $_{1.39}$ & 75.82 & 75.82 & 75.82 \\ 
		Nearest (Gold) & 84.71 $_{0.39}$ & 83.28 $_{1.34}$ & 83.59 $_{0.09}$ & 79.40 & 78.18 & 79.03 \\ \bottomrule
	\end{tabular}
\caption{Results on various embedding models. Right subscript numbers are standard deviations. \textit{embed-ada} refers to text-embedding-ada. \textit{Gold} indicates access to the gold labeled data, thus is not comparable with the rest of methods. Two-stage majority voting is used here. Texts in \textbf{bold} are the best results.}
\label{tab:app_embs}
\end{table*}

\section{Results on Various Number of Demonstrations}
\label{sec:app_demo_num}
We investigate the performance on various number of demonstrations in the input context, the results are in Table \ref{tab:app_demo_num}. As shown in the table, the quantity of examples is not always proportional to the final performance. Similar findings have also been mentioned in \citet{min-etal-2022-rethinking}. We hypothesize that after the LLM learns the mapping between the input-output examples, new information gained from more examples is marginal and might be offset by the more noise introduced.
\begin{table*}[t]
	\centering
	\small
	\begin{tabular}{l c c c c c c}
		\toprule
        \textbf{Numbe of demonstrations} & \textbf{0} & \textbf{2} & \textbf{4} & \textbf{8} & \textbf{16} & \textbf{32} \\ \midrule
		\textit{WikiGold} & ~ & ~ & ~ & ~ & ~ & ~ \\
		Random & 70.80  & 70.25  & 70.86  & \textbf{71.74}  & 71.39  & 70.35  \\ 
		Nearest & 70.80  & 70.41  & 71.32  & 70.47  & \textbf{72.57}  & 71.81  \\ 
		Random (Gold) & 70.80  & 71.75  & 71.54  & \textbf{75.79}  & 73.95  & 74.43  \\ 
		Nearest (Gold) & 70.80  & 76.14  & 77.66  & \textbf{78.97}  & 78.34  & 77.05  \\
		\midrule
		\textit{CoNLL03} & ~ & ~ & ~ & ~ & ~ & ~ \\ 
		Random & 68.97  & 69.54  & 70.84  & 70.53  & 70.72  & \textbf{71.95}  \\ 
		Nearest & 68.97  & 70.12  & 69.15  & 70.90  & 71.81  & \textbf{72.44}  \\ 
		Random (Gold) & 68.97  & 71.94  & 72.76  & 75.12  & 77.81  & \textbf{80.43}  \\ 
		Nearest (Gold) & 68.97  & 79.07  & 80.81  & 83.20  & \textbf{84.12}  & 83.94 \\
		\bottomrule
	\end{tabular}
\caption{Results on various number of demonstrations in the input context. \textit{Gold} indicates access to the gold labeled data, thus is not comparable with the rest of methods. Two-stage majority voting is used here. Texts in \textbf{bold} are the best results. Since the standard deviation values of CoNLL03 are around the same level as in Table \ref{tab:main_results}, we omit them here.}
\label{tab:app_demo_num}
\end{table*}

\section{More Results on Threshold Filtering}
\label{sec:app_various_th}

Table \ref{tab:results_various_thresholds} shows the results of various values of entity-level and sample-level SC thresholds.

\section{Upper Bound of Reliable Annotation Selection}
\label{sec:app_SC_analy}

Table \ref{tab:comlete_SC_analy_results} summarizes the complete results of the upper bound of reliable annotation selection.

\begin{comment}
\begin{figure}[!ht]
	\centerline{\includegraphics[width=0.8\linewidth]{figs/ulb_size_ace05.png}}
	\caption{Increasing unlabeled data on ACE05.}
	\label{fig:ulb_size_ace05}
	\vspace{-1em}
\end{figure}

\begin{figure}[!ht]
	\centerline{\includegraphics[width=0.8\linewidth]{figs/ulb_size_genia.png}}
	\caption{Increasing unlabeled data on ACE05.}
	\label{fig:ulb_size_genia}
	\vspace{-1em}
\end{figure}
\end{comment}

\section{Illustration of Iterative Self-improving}
\label{sec:app_iterations}

\begin{figure*}[t]
	\centerline{\includegraphics[width=0.7\linewidth]{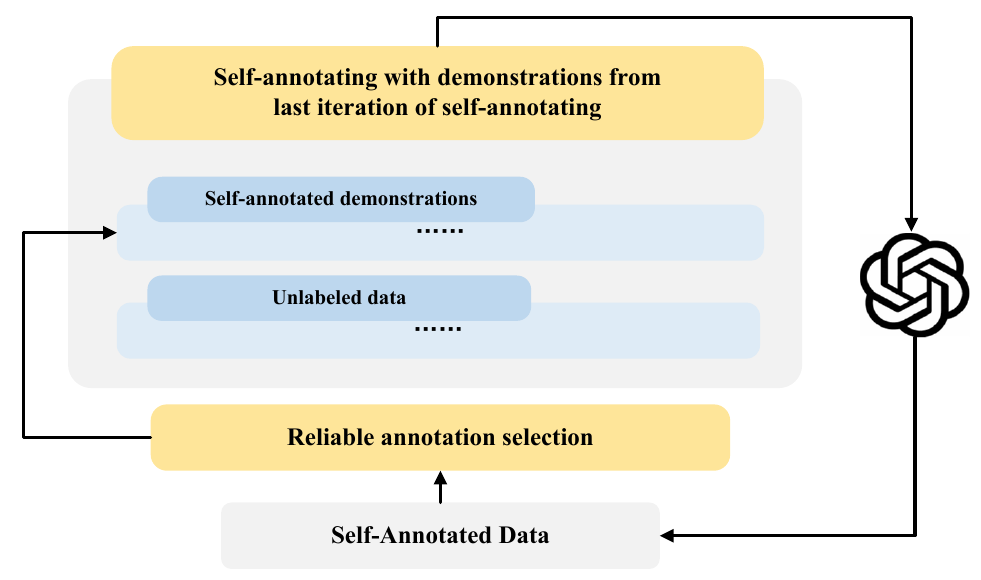}}
	\caption{The pipeline of iterative self-improving.}
	\label{fig:iteration_method}
	\vspace{+1em}
\end{figure*}

The bootstrapping process of iterative self-improving is shown in Fig. \ref{fig:iteration_method}.

\section{Case Study}
\label{sec:case_study}
\begin{figure*}[h]
	\centerline{\includegraphics[width=0.8\linewidth]{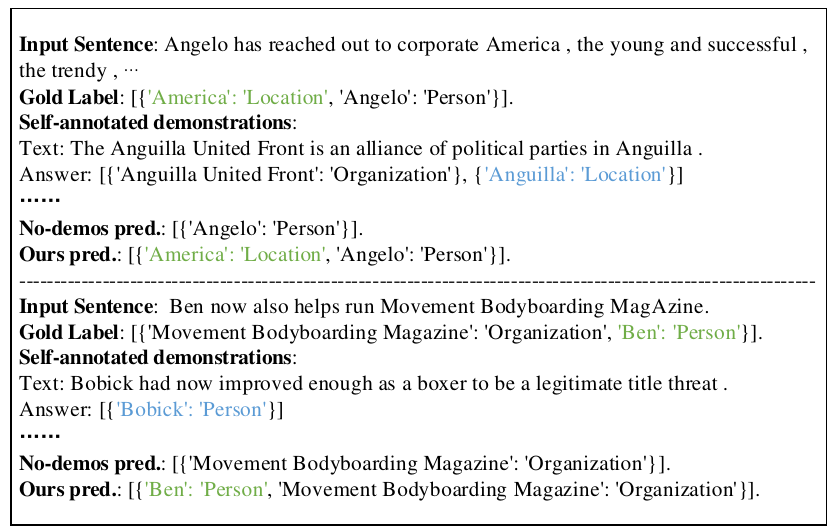}}
	\caption{Case study of self-improving. Examples from WikiGold are illustrated. The errors in predictions of \textit{No-demos} are corrected by our framework with self-annotated demonstrations.  Texts in green are entities corrected by our method. Texts in blue are entities in demonstrations that potentially help with the error correction.}
	\label{fig:case_study}
\end{figure*}

We take a closer look at the cases where the errors in predictions are corrected with self-annotated demonstrations, as shown in Fig. \ref{fig:case_study}. The proposed framework makes the model reuse its own knowledge and correct its own errors, forming a process of self-improving.

%\section{Results on Llama2-13B}
%\label{sec:app_llama2_13b}
%
%To explore the performance of the proposed self-improving framework on weaker LLMs, we conduct experiments on the Llama2 chat 13B model, the results are shown in Table \ref{tab:llama2_13b}. Two-stage majority voting selection strategy and the nearest neighbor retrieval method are used in this experiment. With a much weaker ability in zero-shot scenarios, Llama2 13B model shows negative results under the self-improving framework. This indicates that the proposed framework is more suitable for models with a strong zero-shot capability. For the models with a relatively weaker zero-shot ability, improving the prompt designing might be a more effective strategy to boost performance.
%
%\begin{table}[t]
%	\centering
%	\small
%	%	\tabcolsep=0.2em
%	\begin{tabular}{l c c}
%		\toprule
%		\textbf{Method} & \textbf{CoNLL03} & \textbf{WikiGold} \\ \hline
%		No-demos & 42.24 & 28.57  \\ \midrule
%		TSMV & 23.55 & 8.94  \\ \bottomrule
%	\end{tabular}
%	\caption{Results on the Llama2 chat 13B model. \textit{TSMV} refers to the two-stage majority voting selection strategy, and the nearest retrieval is used. The negative results show that the proposed self-improving framework is more suitable for models with a strong zero-shot capability.}
%	\label{tab:llama2_13b}
%\end{table}

\section{Prompts}
\label{sec:app_prompts}

We show the prompts use in this work in Table \ref{tab:app_prompt_ace05}. We take samples from ACE05 for demonstrations.

\begin{table*}[t]
	\centering
	\small
	\begin{tabular}{l c c c c c}
		\toprule
		\textbf{Method} & \textbf{CoNLL03} & \textbf{ACE05} & \textbf{WikiGold} & \textbf{GENIA} & \textbf{Avg} \\ \midrule
		No-demos & 68.97 $_{0.22}$ & 27.29 $_{0.58}$ & 70.8 & 47.41 $_{0.29}$ & 53.62 \\ \midrule
		\textit{Entity-level SC threshold = 3.0} & ~ & ~ & ~ & ~ & ~ \\ 
		Random & 71.17 $_{0.13}$ & 30.16 $_{0.66}$ & 71.79 & 50.41 $_{0.00}$ & 55.88 \\ 
		Nearest & 71.41 $_{0.66}$ & 31.58 $_{0.76}$ & 73.16 & 51.24 $_{1.79}$ & 56.85 \\ 
		Diverse Nearest, random & 72.68 $_{1.31}$ & 31.39 $_{1.62}$ & 72.01 & 50.65 $_{0.11}$ & 56.68 \\ 
		Diverse Nearest, SC ranking & 73.68 $_{0.03}$ & 31.86 $_{0.13}$ & 73.36 & 51.15 $_{0.69}$ & 57.51 \\ 
		\midrule
		\textit{Entity-level SC threshold = 4.0} & ~ & ~ & ~ & ~ & ~ \\ 
		Random & 70.91 $_{0.55}$ & 30.41 $_{0.95}$ & 72.33 & 50.70 $_{1.53}$ & 56.09 \\ 
		Nearest & 73.24 $_{0.53}$ & 32.22 $_{0.38}$ & 72.53 & 49.85 $_{1.20}$ & 56.96 \\ 
		Diverse Nearest, random & 74.11 $_{0.12}$ & 32.29 $_{0.31}$ & 72.01 & 50.68 $_{0.14}$ & 57.27 \\ 
		Diverse Nearest, SC ranking & 74.99 $_{0.20}$ & 31.65 $_{0.97}$ & 73.53 & 51.11 $_{0.28}$ & 57.82 \\
		\midrule
		\textit{Entity-level SC threshold = 5.0} & ~ & ~ & ~ & ~ & ~ \\ 
		Random & 72.53 $_{0.07}$ & 29.44 $_{0.73}$ & 72.13 & 50.65 $_{0.57}$ & 56.18 \\ 
		Nearest & 74.24 $_{0.03}$ & 29.65 $_{1.30}$ & 72.45 & 48.12 $_{0.45}$ & 56.11 \\ 
		Diverse Nearest, random & 73.50 $_{0.14}$ & 30.55 $_{0.27}$ & 71.34 & 49.34 $_{0.27}$ & 56.18 \\ 
		Diverse Nearest, SC ranking & 72.50 $_{0.66}$ & 30.14 $_{0.35}$ & 74.01 & 49.57 $_{0.61}$ & 56.55 \\
		\midrule
		\textit{Sample-level SC threshold = 3.0} & ~ & ~ & ~ & ~ & ~ \\ 
		Random & 70.17 $_{0.00}$ & 28.78 $_{1.71}$ & 71.81 & 50.45 $_{0.34}$ & 55.30 \\ 
		Nearest & 69.48 $_{0.90}$ & 30.39 $_{0.17}$ & 70.33 & 51.76 $_{0.29}$ & 55.49 \\ 
		Diverse Nearest, random & 68.98 $_{0.86}$ & 30.04 $_{0.34}$ & 69.71 & 51.71 $_{1.41}$ & 55.11 \\ 
		Diverse Nearest, SC ranking & 74.32 $_{1.37}$ & 30.73 $_{0.04}$ & 74.44 & 52.31 $_{0.34}$ & 57.95 \\ 
		\midrule
		\textit{Sample-level SC threshold = 4.0} & ~ & ~ & ~ & ~ & ~ \\ 
		Random & 72.41 $_{1.28}$ & 30.05 $_{1.26}$ & 73.38 & 51.61 $_{1.21}$ & 56.86 \\ 
		Nearest & 72.28 $_{0.14}$ & 32.00 $_{0.08}$ & 73.27 & 52.72 $_{0.80}$ & 57.57 \\ 
		Diverse Nearest, random & 72.32 $_{0.08}$ & 30.74 $_{0.06}$ & 72.09 & 52.50 $_{0.50}$ & 56.91 \\ 
		Diverse Nearest, SC ranking & 73.97 $_{0.12}$ & 31.08 $_{0.54}$ & 72.80 & 51.67 $_{0.93}$ & 57.38 \\ 
		\midrule
		\textit{Sample-level SC threshold = 5.0} & ~ & ~ & ~ & ~ & ~ \\ 
		Random & 73.66 $_{0.69}$ & 29.19 $_{0.26}$ & 71.92 & 51.34 $_{0.97}$ & 56.52 \\ 
		Nearest & 74.19 $_{0.30}$ & 30.94 $_{0.11}$ & 74.96 & 52.01 $_{0.23}$ & 58.02 \\ 
		Diverse Nearest, random & 73.16 $_{0.66}$ & 27.98 $_{0.08}$ & 74.55 & 50.64 $_{0.18}$ & 56.58 \\ 
		Diverse Nearest, SC ranking & 74.53 $_{0.51}$ & 30.00 $_{0.73}$ & 73.60 & 51.02 $_{0.98}$ & 57.28 \\ 
		\bottomrule
	\end{tabular}
	\caption{Results of various entity-level SC thresholds and sample-level SC thresholds. Right subscript numbers are standard deviations.}
	\label{tab:results_various_thresholds}
\end{table*}

% braket
%\begin{table*}[t]
%	\centering
%	\small
%	\begin{tabular}{l l l l l l}
	%		\toprule
	%		\textbf{Method} & \textbf{CoNLL03} & \textbf{ACE05} & \textbf{WikiGold} & \textbf{GENIA} & \textbf{Avg} \\ \midrule
	%		No-demos & 68.97 (0.22) & 27.29 (0.58) & 70.8 & 47.41 (0.29) & 53.62 \\ \midrule
	%		\textit{No SC} & ~ & ~ & ~ & ~ & ~ \\ 
	%		Random & 71.45 (0.10) & 30.38 (0.93) & 70.51 & 48.78 (0.06) & 55.28 \\ 
	%		Nearest & 72.07 (0.11) & 32.20 (0.92) & 71.81 & 49.54 (1.88) & 56.40 \\ 
	%		Diverse Nearest, random & 72.15 (0.65) & 31.07 (1.45) & 70.72 & 50.01 (1.20) & 55.99 \\
	%		\midrule
	%		\textit{Two-stage majority voting} & ~ & ~ & ~ & ~ & ~ \\ 
	%		Random & 72.12 (0.59) & 31.18 (0.38) & 72.32 & 50.17 (0.93) & 56.45 \\ 
	%		Nearest & 71.66 (0.37) & 31.45 (1.32) & 72.84 & 50.19 (1.59) & 56.53 \\ 
	%		Diverse Nearest, random & 72.45 (0.41) & 30.84 (0.56) & 70.83 & 51.03 (0.73) & 56.28 \\ 
	%		Diverse Nearest, SC ranking & 74.51 (0.03) & 32.27 (0.25) & 73.98 & 52.06 (0.09) & 58.20 \\
	%		\midrule
	%		\textit{True answer of SC} & ~ & ~ & ~ & ~ & ~ \\ 
	%		Random & 73.72 (0.41) & 32.71 (0.56) & 73.83 & 52.67 (0.09) & 58.23 \\ 
	%		Nearest & 81.65 (0.17) & 37.82 (0.59) & 76.57 & 56.24 (0.44) & 63.07 \\ 
	%		Diverse Nearest, random & 78.84 (1.43) & 35.79 (0.26) & 76.20 & 54.46 (0.98) & 61.32 \\ 
	%		Diverse Nearest, SC ranking & 80.12 (0.02) & 35.23 (0.63) & 76.64 & 54.58 (0.57) & 61.64 \\
	%		\bottomrule
	%	\end{tabular}
%	\caption{Complete results of removing SC (No SC) and potential upper bound of SC (True only).}
%	\label{tab:comlete_SC_analy_results}
%\end{table*}

% subscript
\begin{table*}[t]
	\centering
	\small
	\begin{tabular}{l c c c c c}
		\toprule
		\textbf{Method} & \textbf{CoNLL03} & \textbf{ACE05} & \textbf{WikiGold} & \textbf{GENIA} & \textbf{Avg} \\ \midrule
		No-demos & 68.97 $_{0.22}$ & 27.29 $_{0.58}$ & 70.8 & 47.41 $_{0.29}$ & 53.62 \\ \midrule
		\textit{Two-stage majority voting} & ~ & ~ & ~ & ~ & ~ \\ 
		Random & 72.12 $_{0.59}$ & 31.18 $_{0.38}$ & 72.32 & 50.17 $_{0.93}$ & 56.45 \\ 
		Nearest & 71.66 $_{0.37}$ & 31.45 $_{1.32}$ & 72.84 & 50.19 $_{1.59}$ & 56.53 \\ 
		Diverse Nearest, random & 72.45 $_{0.41}$ & 30.84 $_{(0.56}$ & 70.83 & 51.03 $_{0.73}$ & 56.28 \\ 
		Diverse Nearest, SC ranking & 74.51 $_{0.03}$ & 32.27 $_{0.25}$ & 73.98 & 52.06 $_{0.09}$ & 58.20 \\
		\midrule
		\textit{Upper bound} & ~ & ~ & ~ & ~ & ~ \\ 
		Random & 73.72 $_{0.41}$ & 32.71 $_{0.56}$ & 73.83 & 52.67 $_{0.09}$ & 58.23 \\ 
		Nearest & 81.65 $_{0.17}$ & 37.82 $_{0.59}$ & 76.57 & 56.24 $_{0.44}$ & 63.07 \\ 
		Diverse Nearest, random & 78.84 $_{1.43}$ & 35.79 $_{0.26}$ & 76.20 & 54.46 $_{0.98}$ & 61.32 \\ 
		Diverse Nearest, SC ranking & 80.12 $_{0.02}$ & 35.23 $_{0.63}$ & 76.64 & 54.58 $_{0.57}$ & 61.64 \\
		\bottomrule
	\end{tabular}
	\caption{Complete results of the upper bound of reliable annotation selection. Right subscript numbers are standard deviations.}
	\label{tab:comlete_SC_analy_results}
\end{table*}

\begin{table*}[t]
	\centering
	\begin{tabular}{l}
		\toprule
		Prompts of zero-shot setting\\ 
		\midrule
		Given entity label set: {[}'Person', 'Organization', 'Location', 'Facility', 'Weapon',\\ 'Vehicle', 'Geo-Political Entity'{]}.\\
		Please recognize the named entities in the given text.  Based on the given entity label set, \\
		$ $provide answer in the following JSON format: [\{'Entity Name': 'Entity Label'\}]. If there\\
		$ $ is no entity in the text, return the following empty list: []. \\
		\\
		Text: right now we 're also waiting to hear from the president at the white house .\\
		Answer: \\
		\midrule
		
		Prompts of ICL \\ 
		\midrule
		Given entity label set: {[}'Person', 'Organization', 'Location', 'Facility', 'Weapon',\\ 'Vehicle', 'Geo-Political Entity'{]}.\\
		Please recognize the named entities in the given text.  Based on the given entity label set, \\
		$ $provide answer in the following JSON format: [\{'Entity Name': 'Entity Label'\}]. If there\\
		$ $ is no entity in the text, return the following empty list: []. \\
		\\
		Text: right now we 're also waiting to hear from the president at the white house .\\
		Answer: [\{'white house': 'Location'\}, \{'president': 'Person'\}]\\
		\\
		Text: At the Pentagon , Barbara Starr reports officials say today begins a new strategy\\
		 in the skies over Baghdad .\\
		Answer: [\{'Barbara Starr': 'Person'\}, \{'Pentagon': 'Facility'\}, \{'officials': 'Person'\},\\ \{'skies':'Location'\}, \{'Baghdad': 'Geo-Political Entity'\}]\\
		\\
		Text: John Irvine , ITV News , Baghdad .\\
		Answer: [\{'John Irvine': 'Person'\}, \{'ITV News': 'Organization'\}, \\
		\{'Baghdad': Geo-Political Entity'\}]\\
		... ...\\
		\\
		Text: right now we 're also waiting to hear from the president at the white house .\\
		Answer: \\
        \\ 
        \bottomrule
	\end{tabular}
	\caption{Prompts used in this work. A few samples from ACE05 are displayed for demonstrations.}
	\label{tab:app_prompt_ace05}
\end{table*}

\end{document}